\pdfoutput=1

\documentclass[11pt]{article}

\usepackage[preprint]{acl}

\usepackage{times}
\usepackage{latexsym}

\usepackage{booktabs}
\usepackage{url}
\usepackage{hyperref}
\usepackage{adjustbox}
\usepackage{multirow}
\usepackage{float}
\usepackage{subcaption}
\usepackage{tcolorbox}

\newtcolorbox{frabox}[1]{colback=orange!5!white,colframe=pink!75!black,title=#1}

\newtcolorbox{sptbox}[1]{colback=green!5!white,colframe=yellow!75!black,title=#1}

\newtcolorbox{combox}[1]{colback=violet!5!white,colframe=magenta!75!black,title=#1}

\usepackage[T1]{fontenc}

\usepackage[utf8]{inputenc}

\usepackage{microtype}

\usepackage{inconsolata}

\usepackage{graphicx}

\title{Do Construction Distributions Shape Formal Language Learning In German BabyLMs?}

\author{Bastian Bunzeck \and Daniel Duran \and Sina Zarrieß \\
  CRC 1646 -- Linguistic Creativity in Communication \\
  Department of Linguistics \\
  Bielefeld University, Germany \\
  \texttt{\{firstname.lastname\}@uni-bielefeld.de}}

\begin{document}
\maketitle
\begin{abstract}
We analyze the influence of utterance-level construction distributions in German child-directed/child-available speech on the resulting word-level, syntactic and semantic competence (and their underlying learning trajectories) in small LMs, which we train on a novel collection of developmentally plausible language data for German. We find that trajectories are surprisingly robust for markedly different distributions of constructions in the training data, which have little effect on final accuracies and almost no effect on global learning trajectories. While syntax learning benefits from more complex utterances, word-level learning culminates in better scores with more fragmentary utterances. We argue that LMs trained on developmentally plausible data can contribute to debates on how conducive different kinds of linguistic stimuli are to language learning.
\end{abstract}


\section{Introduction}

One of the most contentious issues in language acquisition is the relationship between the input that learners receive and the resulting linguistic system \cite{pullum2002empirical,clark2011linguistic}. Child-directed speech (or \textit{CDS}) is structurally simple: Especially in the first three years of life, it abounds with questions, imperatives, and fragmentary utterances, but features fewer SV(X) and very few complex sentences, which instantiate ``canonical'' word order \cite{cameron-faulkner2003construction}. This distribution of utterance-level constructions is conducive to the \textit{functional} side of language acquisition: caregivers talk in this way to elicit responses, steer behavior, or establish joint attention. But how do children acquire full-fledged, \textit{formal} grammatical knowledge from such supposedly skewed input? While its advantages for aspects like speech segmentation or word learning are somewhat accepted \cite{yurovsky2012statistical,cristia2019segmentability}, its influence on syntax remains debated: whereas some generativist approaches see any kind of input as too impoverished to learn a full-fledged syntactic system (cf. \citealp{chomsky1965aspects,crain2001nature,guasti2002language,thomas2002development,berwick2011poverty}), constructionist and usage-based scholars argue that this supposedly skewed input actually aids syntax learning \cite{macwhinney2004multiple,tomasello2005formalities,bunzeck2024richness}.

\begin{figure}[t]
\centering
\footnotesize
\begin{combox}{\textbf{Project Gutenberg:} Complex sentences}
Aber sie war in Angst, dass wir die Larven beschädigen würden, die zu Arbeiterinnen heranwachsen sollten. (\textit{But she was afraid that we would damage the larvae which were supposed to grow into workers.})
\end{combox}
\begin{sptbox}{\textbf{MiniKlexikon:} Transitive SP sentences}
Der Grafiker entwirft das Bild vorne auf dem Buch. (\textit{The graphic designer designs the picture on the cover of the book.}) \\
Der Friseur schneidet die Haare. (\textit{The hairdresser cuts the hair.})
\end{sptbox}
\begin{frabox}{\textbf{Child-directed speech:} Fragmentary utterances}
noch mehr! (\textit{even more!}) \\
ja. (\textit{yes.}) \\
mit dem Flugzeug. (\textit{with the airplane.})
\end{frabox}
\vspace{-0.3cm} 
\caption{Examples for most frequent construction types from different portions of our German BabyLM corpus}
\vspace{-0.6cm} 
\label{fig:cxn-teaser}
\end{figure}


The connectionist ``renaissance'', fueled by deep learning and Transformer language models, has opened up new avenues of investigating the relationship between an artificial learner's acquired linguistic system and the nature of its training data, more recently also from a constructionist/non-generativist viewpoint \cite{weissweiler2023construction,piantadosi2024modern}. LLMs, pretrained on raw language data only, and instruction-finetuned chatbots based on them, generate text without grammatical errors, and perform well in controlled syntactic test suites. Unfortunately, though, their massive parameter size does not preclude the possibility that their linguistic capabilities result from memorization rather than generalization \cite{milliere2024language}. Furthermore, the sheer amount of their pretraining data exceeds human learner's input by many orders of magnitude, putting their relevance for linguistic modeling into question. Work within the BabyLM community \cite{warstadt2023findings, hu2024findings, charpentier2025babylm} has demonstrated that Transformer LMs, trained on cognitively plausible amounts of data, can often acquire fairly complex syntactic structures, even without instruction-finetuning. They can also learn accurate word-level representations when trained with character-level tokenization \citep{bunzeck2025subword,goriely2025babylms}. This makes them ideal testbeds for the aforementioned issue: does the construction distribution found in CDS, which features a high proportion of questions and syntactic fragments, affect the acquisition of formal linguistic capabilities? In other words, does robust linguistic knowledge at the word and syntax level emerge when the training data is closer to the fragmented, ``messy'' input of human learners?


The goals of this paper, then, are twofold: (1) we compile a novel German BabyLM training set, for which we conduct the first utterance-level construction analysis for German. We find that distributions align with findings for English and other languages. We then (2) create three 5M-token subsets with distinct constructional profiles, varying, e.g., the proportion of fragmentary and complex utterances, and train small, character-based and subword Llama models on them. We evaluate them with lexical, syntactic, and semantic minimal pairs \cite{bunzeck2025small,mueller2020crosslinguistic,he2025xcomps} to gauge the influence of different construction distributions on these levels of linguistic knowledge, and find that differences between grammatically complex training data and a developmentally plausible constructional distribution are fairly small. While certain syntactic phenomena are learned somewhat better from more complex sentences, lexical learning improves with more fragments and questions in the input. Most interestingly, input complexity only modulates the steepness of the resulting learning trajectories, but has no principal effect on the amount of input needed to kickstart learning. 

\section{Constructions in children's input}

Child-directed speech can be seen as a separate linguistic register and is the primary input that children encounter in their first years. On the phonetic level, it features slower speech and exaggerated intonation patterns, which infants prefer listening to \cite{zangl2007increased}, while its vocabulary is mostly restricted to everyday topics and children's immediate surroundings \cite{snow1977talking}. Structurally, child-directed utterances are usually shorter and simpler than adult-directed ones \cite{genovese2020infantdirected} and feature high amounts of structural and lexical repetition \cite{tal2024infantdirected}. Statistical properties of the input directly influence the children's order of acquisition for syntactic patterns \cite{huttenlocher2002language,ambridge2015ubiquity}, e.g., for relative clauses \cite{diessel2000development,brandt2008acquisition,chen2015acquisition}.

Early studies were mostly concerned with mapping out how much CDS is ungrammatical or otherwise ``wrong'' (in the sense of hesitations, false starts, etc., cf. \citealp{pine1994language}), but the quantitative turn in linguistics \cite{janda2013cognitive} has enabled more holistic analyses. In a seminal study, \citet{cameron-faulkner2003construction} analyze utterance-level constructions in child-directed English via corpora of toyplay sessions featuring children and caregivers. They show that CDS features only few ``canonical'' SV(X)-utterances but abounds with questions, lexical fragments, or copula constructions. The reported construction distributions also hold for typologically different languages, e.g., Irish \cite{cameron-faulkner2011form}.
These constructions and their real-world functions help children to quickly understand the \textit{functional} side of language. However, the most common and repetitive utterances that English-speaking children hear represent a rather skewed sample of the presumed, underlying \textit{formal} language system. Generativist approaches would argue that certain formal processes, like question formation from relative clauses, are not attainable from this kind of language, as the input never contains specific examples \cite{chomsky1980rules} (although \citealp{pullum2002empirical} find that the input frequently contains exactly such specific examples). They also partly emphasize the importance of statistical learning, e.g. for providing hypotheses about competing possible mental grammars constrained by innate, language-specific mechanisms (cf. \citealp{yang2004universal}, also \citealp[121f.]{ambridge2011child}). Constructivist approaches do not view language learning as such a re-construction of the target language's abstract grammar, but rather as the re-construction of the target language's inventory of form-meaning pairings \cite{behrens2021constructivist}. They argue that this kind of input is actually conducive to formal aspects of acquisition, by providing anchor points for first words and their semantic links to real-world reference, which then serve as building blocks for a gradual development into larger schemas (like questions with relative clauses).

Although CDS features such a skewed construction distribution, written language aimed at children, e.g., in children's books, is characterized by a much higher rate of canonical SV(X)-constructions than CDS \cite{cameron-faulkner2013comparison}. Questions rarely occur in books. CDS produced in shared book reading presents a middle-ground --- it contains more complex and SV(X)-constructions than regular CDS, but less than book text alone \cite{noble2018keeping}. They argue that shared reading therefore, plays an important role in moving children from early, isolated traces of linguistic knowledge to a rich mental language system. This also aligns with the findings by \citet{bunzeck2024richness}, who show that the distribution of constructions in CDS varies with situation type (toyplay features most questions, meal sessions beget more imperatives, shared book reading features more complex constructions) and child age (questions and imperatives become less frequent with age). They suggest that CDS is therefore adapted to support children's cognitive and linguistic development. Yet, as corpus studies are necessarily descriptive and cannot establish causal/mechanistic connections on their own (e.g. what would happen if a child never hears CDS), it remains questionable if this is actually true. Here, the potential of LMs trained on little data  becomes apparent for constructionist approaches: they allow controlled experiments with different kinds of input data, which can serve as additional evidence for effects hypothesized from corpus data.

\section{Input in developmentally plausible LMs}

\paragraph{Authentic data}
Early approaches to modeling language acquisition with neural networks used hand-picked, manually ordered data points \cite{rumelhart1986learning} or synthetic data generated with hand-crafted grammars \cite{elman1993learning,christiansen1999connectionist,chang2006becoming}. Both lack developmental plausibility. Since then, data availability has improved with the establishment of developmental corpora. Frequently, CDS from CHILDES  \cite{macwhinney2000childes} is used to train developmentally plausible LMs (cf. \citealp{pannitto2020recurrent,huebner2021babyberta}). While CHILDES-based models have the advantage of learning from authentic data \textit{only}, they have the disadvantage of not accessing the \textit{full breadth} of the linguistic input children receive. Children are exposed to many more different registers of language throughout their linguistic development, like shared (or solitary) book reading, or television shows \cite{montag2019differences,gowenlock2024exposure}. In response to this, the BabyLM corpora propose a data mix of varied spoken and written sources, from CDS over adult-adult conversations to OpenSubtitles \cite{lison2016opensubtitles2016}, but also children's \cite{hill2015goldilocks} and adults' books \cite{gerlach2020standardized}. All data included in them could be plausibly encountered by children, which provides opportunities to ablate the influence of architecture/training on the learned linguistic knowledge.

\begin{table*}[htbp]
\footnotesize
\centering
\begin{tabular}{@{}llr@{}}
\toprule
Dataset & Description & \# Words \\ \midrule
\multirow{2}{*}{CHILDES \cite{macwhinney2000childes}} & Child-directed speech & 3,626,301 \\
 & Child speech & 1,511,144 \\
OpenSubtitles \cite{lison2016opensubtitles2016} & Movie subtitles & 1,543,094 \\
CallHome \cite{karins1997callhome} & Phone conversations & 176,313 \\
Klexikon & Children's online encyclopedia & 1,384,891 \\
MiniKlexikon & Simplified online encyclopedia & 272,886 \\
Wikibooks Wikijunior & Educational books & 226,773 \\
Fluter & German youth magazine & 2,862,278 \\
Project Gutenberg & Literature (children's and young adult) & 2,476,133 \\
Dreambank \cite{domhoff2008studying} & Dream reports & 939,197 \\
Leipzig corpus news texts \cite{goldhahn2012building} & Short news texts & 1,541,803 \\ \midrule
\textit{Total} & & 16,560,813 \\ \bottomrule
\end{tabular}
\vspace{-0.1cm}
\caption{Lexical token counts for all subcorpora of our corpus}
\vspace{-0.4cm}
\label{tab:german-babylm}
\end{table*}

For languages other than English, data availability is the greatest problem for the construction of developmentally plausible datasets. \citet{salhan2024less} and \citet{padovani2025childdirected} use only data available from CHILDES for models in different languages, whereas \citet{prevot2024extending} compare models trained on spoken data (child-directed + adult-adult conversations) with models trained on the French Wikipedia. As such, these first forays into more polyglot BabyLMs are still constrained to the child-directed input found in CHILDES and do not extend to the aforementioned variety of inputs  \cite{soderstrom2007babytalk,gowenlock2024exposure}. Notably, \citet{suozzi2025bambi} introduce an Italian BabyLM but do not elaborate on their data sources beyond CHILDES.

\paragraph{Linguistic properties}
The linguistic make-up of pre-training data and its influence on linguistic performance have only recently begun to receive increased scrutiny. Focusing on the lexical level, \citet{yam2024what} measure sentence-level textual complexity with readability metrics based on text-wide word/syllable--sentence ratios for different corpora (CHILDES, BabyLM corpus, synthetic data, Project Gutenberg). They find that models trained on more complex text perform better at syntactic benchmarks, but simpler data (CHILDES) is learned better in terms of perplexity and loss convergence. \citet{muckatira2024emergent} filter English pre-training corpora for text spans that only contain vocabulary also found in English CHILDES data and find that simplified models generate more coherent text than models trained on more complex data, and also succeed in syntactic tests if the test data is filtered accordingly. In contrast, \citet{edman2024are} change the semantic content of the pre-training data and use datasets that approximate the linguistic input second-language learners get, e.g., dictionary entries, grammar books, and paraphrases. While grammar books moderately improve syntactic evaluation, there is no positive effect for the addition of the other text types. 

\paragraph{Filtered corpora} 
While actual research on the syntactic properties of the input is rather rare, training on filtered corpora has been used in pilot studies. \citet{patil2024filtered} and \citet{misra2024language} filter out specific grammatical constructions from the BabyLM corpora and then probe the resulting models for knowledge of these grammatical constructions (which might also be analogically learned from related constructions or constructed from their parts). \citet{patil2024filtered} show that their models succeed on the BLiMP benchmark \cite{warstadt2020blimp}, even if sentences containing structures targeted in BLiMP's minimal pair sets are removed. Similarly, \citet{misra2024language} show that acceptability scores for the English AANN construction can be reliably estimated from models that have never seen it. In sum, then, models appear to be able to generalize from indirect evidence and learn language in a somewhat constructivist, bottom-up fashion.

The structural composition of child-directed data has (so far) not been scrutinized. Most studies focus on lexical or semantic properties, emphasizing content over structure; child-directed data is usually equated with a somewhat fitting vocabulary or with just being authentic data. However, findings from usage-based linguistics suggest that structural properties, like utterance-level construction distributions, play a crucial role in language acquisition. Whereas \citet{patil2024filtered} and \citet{misra2024language} remove specific constructions from their data, we aim to explore whether different global distributions of constructions influence the resulting linguistic knowledge and learning trajectories.

\section{A German BabyLM dataset}

\begin{figure*}[ht!]
\centering
\includegraphics[width=0.97\linewidth]{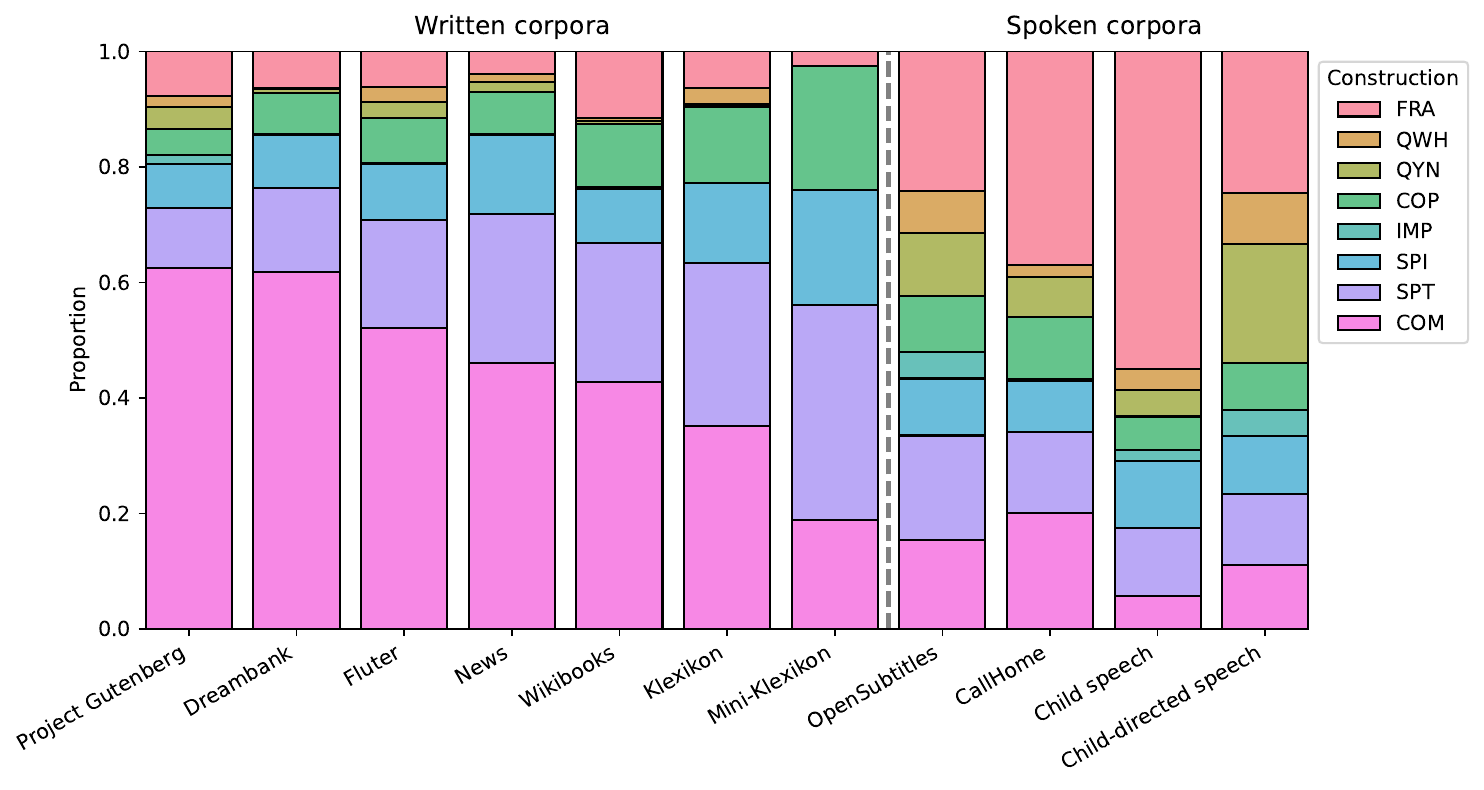}
\vspace{-0.4cm}
\caption{Proportions of utterance-level constructions for all subcorpora in our corpus}
\label{fig:cxn-proportions}
\vspace{-0.3cm}
\end{figure*}

To construct a German dataset, we use a variety of developmentally plausible sources, similar to the English BabyLM data \cite{warstadt2023findings,choshen2024call}. We use (1) all data from German CHILDES corpora \cite{macwhinney2000childes}, including frog stories from TalkBank \cite{berman1994different} and math lessons from ClassBank \cite{stigler2000using}, (2) subtitles from OpenSubtitles \cite{lison2016opensubtitles2016}, (3) adult conversations from the CallHome corpus \cite{karins1997callhome}, and (4) written data from Project Gutenberg, from which we downloaded a manually curated sample of children's books, young adult literature and literature commonly read in German schools. We supply this data with two corpora, the DreamBank database of self-reported dreams \cite{domhoff2008studying} and short news texts from the Leipzig corpus \cite{goldhahn2012building}; although they are not child-directed per se, these sources are child-available in everyday language. 

To approximate child-available input even better, we tap into freely available child/learner-directed sources and compile four additional subcorpora for our dataset. The Wikibooks Wikijunior shelve features educational resources aimed at children, focusing on a diverse array of topics such as technology or nature. The Klexikon is a children's wiki in German, featuring more than 3,000 articles aimed at children between 5--15. A simplified version of it is the MiniKlexikon, which features over 1,500 articles aimed at beginning readers. Finally, we also scrape the complete archives of \textit{Fluter}, a magazine aimed at young adults published by the Federal Agency for Civic Education, which contains a large body of non-fiction. All resources are CC-licensed. Table \ref{tab:german-babylm} shows the raw token numbers for all corpora (16.5M overall). We extensively clean and normalize our data (details in Appendix \ref{app:data-cleaning}) and make our dataset available on Hugging Face.\footnote{\url{https://huggingface.co/datasets/bbunzeck/babylm-german}}

\section{Construction distribution analysis}

As there are no findings on the distribution of utterance-level construction in German, we conduct our own analysis using \texttt{spacy} \cite{honnibal2020spacy}. We first split larger paragraphs into individual sentences with the included senter and then annotate these with POS and dependency information. This information serves as the base of our construction annotation procedure. We devise standard construction categories in line with comparable efforts for English \cite{cameron-faulkner2003construction,cameron-faulkner2013comparison,bunzeck2024richness}, and assign one of the following categories to each utterance:

\begin{itemize}
    \setlength\itemsep{-0.4em}
    \item \textbf{FRA} -- utterances that do not contain a verb 
    \item \textbf{QWH} -- wh-question (introduced by interrogative pronouns)
    \item \textbf{QYN} -- yes/no-question (introduced by verbs/auxiliaries)
    \item \textbf{COP} -- subject-predicate utterance where the predicate is a copula verb (a form of \textit{sein} or \textit{werden})
    \item \textbf{IMP} -- utterances introduced by verbs in imperative mood
    \item \textbf{SPI} -- standard subject-predicate utterance (intransitive verb with no direct/accusative object)
    \item \textbf{SPT} -- standard subject-predicate utterance (transitive verb with direct/accusative object)
    \item \textbf{COM} -- utterances with two or more lexical verbs    
\end{itemize}

This holistic taxonomy is applicable to every utterance in our corpus. For a balanced, manually annotated sample of 1,000 sentences our classifier reaches an accuracy of approx. 95\%.

Figure \ref{fig:cxn-proportions} visualizes the results of this annotation process, exact proportions are reproduced in Appendix \ref{app:exact-cxn-proportions}. Generally, our results confirm earlier findings \citep{cameron-faulkner2003construction,cameron-faulkner2011form,cameron-faulkner2013comparison,bunzeck2024richness}: Just like English CDS, German CDS features more questions than any other corpus, abounds with fragments, and contains comparatively few complex utterances. The Project Gutenberg data, on the other hand, is characterized by over 60\% complex sentences. Interestingly, the construction distribution forms a continuum across our subcorpora. The MiniKlexikon, for example, contains considerably less complex sentences than the other written genres, but over half of its utterances are (in)transitive, canonical SV-sentences. This shows that even these particular sub-genres of child-directed linguistic input feature highly varied and specific constructional profiles that differ from each other.

\section{Training data composition}

We compose three different corpora of 5M words: (1) one corpus maximally resembling the construction composition of child-directed speech (\texttt{cds}), (2) one corpus containing a drastically higher amount of complex sentences, mirroring the distribution in the Project Gutenberg data (\texttt{pjg}), and (3) a corpus that is averaged between these two (\texttt{mix}). The relative distributions of construction types can be found in Table \ref{tab:three-sets}.

\begin{table}[htb!]
\centering
\footnotesize
\begin{tabular}{@{}l|rrr@{}}
\toprule
Construction & \texttt{cds} & \texttt{mix} & \texttt{pjg} \\ \midrule
FRA & 25\% & 16.5\% & 8\% \\
QWH & 9\% & 5.5\% & 2\% \\
QYN & 21\% & 12.5\% & 4\% \\
COP & 8\% & 6.5\% & 5\% \\
IMP & 5\% & 3.5\% & 2\% \\
SPI & 10\% & 9\% & 8\% \\
SPT & 12\% & 11\% & 10\% \\
COM & 10\% & 35.5\% & 61\% \\ \bottomrule
\end{tabular}
\vspace{-0.1cm}
\caption{Construction proportions of our training sets}
\vspace{-0.3cm}
\label{tab:three-sets}
\end{table}

Crucially, we sample the individual utterances for our training sets from all subcorpora in our German BabyLM dataset. By doing so, we approximate a similar (if not completely equal) mixture of sources and, therefore also a similar mixture of registers, semantic content, etc. This enables us to isolate the effect of construction distributions in our model's training data, without any interference from the possible differences between the subcorpora.

\begin{table*}[ht!]
\centering
\footnotesize
\begin{tabular}{@{}ll|ccc|ccc|c@{}}
\toprule
 &  & \multicolumn{3}{c|}{Character} & \multicolumn{3}{c|}{Subword} & Llama 3.2 1B \\ \midrule
 &  & \texttt{cds} & \texttt{mix} & \texttt{pjg} & \texttt{cds} & \texttt{mix} & \texttt{pjg} & -- \\ \midrule
\multirow{3}{*}{Word-level} & Lexical decision & 97.4\% & \textbf{97.6\%} & 97.4\% & 84.6\% & 81.9\% & 80.8\% & 69.6\% \\
 & Surprisal & 99.8\% & 99.8\% & \textbf{99.9\%} & 91.5\% & 90.3\% & 90.1\% & 98\%\\
 & AntiSurprisal & 99.3\% & 98.9\% & \textbf{99.7\%} & 76.5\% & 75.4\% & 75.4\% & 87.4\% \\ \midrule
\multirow{7}{*}{Syntax} & Simple Agreement & 90\% & 90\% & \textbf{95.7\%} & 80\% & 84.3\% & 92.1\% & 95.71\% \\
 & Across a Prepositional Phrase & 61.5\% & 65.5\% & 61.8\% & 74.8\% & 73.5\% & \textbf{75.5\%} & 83\% \\
  & Across a Subject Relative Clause & 67.1\% & 66\% & 62.4\% & 78.4\% & 73.7\% & \textbf{97.9\%} & 99.7\% \\
  & Short Verb Phrase Coordination & 69.8\% & 68.8\% & 67.9\% & 82.6\% & 93.5\% & \textbf{99.5\%} & 99.9\%\\
  & Long Verb Phrase Coordination & 53.6\% & 60.6\% & 63\% & 60.6\% & \textbf{78.8\%} & 78\% & 90.5\% \\
 & Across Object Relative Clause & 58.6\% & 54.2\% & 53\% & 64\% & 66.7\% & \textbf{81.6\%} & 86.1\% \\
 & Within Object Relative Clause & 59.8\% & 56.4\% & \textbf{72.5\%} & 55.8\% & 55.7\% & 49.9\% & 61.4\% \\
 \midrule
Semantics & XCOMPS & 51.5\% & 49.1\% & 49.1\% & 51.4\% & 52\% & \textbf{52.3\%} & 58.9\% \\ \bottomrule
\end{tabular}
\vspace{-0.1cm}
\caption{Final evaluation results (accuracies) for all benchmarks}
\vspace{-0.3cm}
\label{tab:results}
\end{table*}

\section{Model training and evaluation}

We train small Llama models \cite{touvron2023llama} with \texttt{transformers} \cite{wolf2020transformers}. To account for the effect of subword tokenization, we compare character-level (3.7M parameters) and subword models (7.7M parameters) for the three datasets. We train all models for one epoch (loss curves and hyperparameters are in Appendix \ref{app:loss-curves}) and share them on Hugging Face.\footnote{\url{https://huggingface.co/collections/bbunzeck/german-babylm-67b868e08ff8782a9814ceaf}} To test the effect of different random initializations and our sampling strategy, we reproduce pre-training for the \texttt{cds} models (see Appendix \ref{app:repeated-runs}).

In line with current best practices to linguistic probing, we use minimal pair datasets to evaluate our LMs' linguistic knowledge in German. The datasets always consist of a correct/grammatical and a matched incorrect/ungrammatical string. We use \texttt{minicons} \cite{misra2022minicons} to score the sentences and evaluate 19 model checkpoints per model (10 for the first 10\% of training, 9 for the remaining 90\%). As an additional ablation, we also evaluate the multilingual Llama 3.2 1B\footnote{\url{https://huggingface.co/meta-llama/Llama-3.2-1B}} on all probing paradigms. Currently, no monolingual German Llama models exist. Therefore, the medium-sized 1B-parameter version of Llama 3.2, which is trained on a considerable amount of German language data, is a useful baseline for expected benchmark scores enabled through a higher model capacity and more training data.

\paragraph{Word-level probing}
Language acquisition first involves learning what words are, i.e. which (sound) sequences map to word-level items in the mental lexicon, before learning how they combine. To gauge this most basic learning step, we adapt the experimental setup from \citet{bunzeck2025small}: We use \texttt{wuggy} \cite{keuleers2010wuggy} to generate 1,000 nonce words (e.g. \textit{promsen}) from existing words (e.g. \textit{bremsen}) and then evaluate how surprised the models are by (1) the words with the context of a prepended white space (lexical decision, \citealp{legodais2017comparing}), (2) the words in a plausible context sequence (surprisal, \citealp{hale2001probabilistic}), and (3) the words randomly inserted into implausible contexts (antisurprisal, \citealp{shafiabadi2025surprisal}). If the model is less surprised by the existing word, we count this as a correct choice in our paradigm. We calculate accuracies over the whole dataset.

\paragraph{Syntactic probing}
For syntactic probing, we use the CLAMS dataset \cite{mueller2020crosslinguistic}, which contains syntactic minimal pairs (grammatical/ungrammatical) for German (e.g. \textit{Die Autoren lachen/*lacht.}). The included seven phenomena all revolve around subject-verb agreement in different contexts (across PPs, relative clauses, with coordination, etc.), resulting in different degrees of difficulty. We score the sentences for their likelihood. We calculate accuracies for correctly rated pairs (grammatical sentence more likely) over the whole dataset.

\paragraph{Semantic probing}
To evaluate our models' semantic knowledge, we use the XCOMPS dataset \cite{he2025xcomps}. It contains conceptual minimal pairs (e.g. \textit{Garnele hat einen Kopf./*Ein Bikini hat einen Kopf.})\footnote{We sample 1,000 MPs with randomized replacement, as the other conditions contain implausible/wrong minimal pairs. Furthermore, the quality of translation is not optimal, as exemplified by the missing determiner in front of \textit{Garnele}.} that test whether LMs have acquired knowledge about conceptual properties of real-world entities. Again, we score the sentences for likelihood and calculate accuracy over the whole dataset.

\begin{figure*}[ht!]
\centering
\begin{subfigure}{.5\textwidth}
  \centering
  \includegraphics[width=0.9\linewidth]{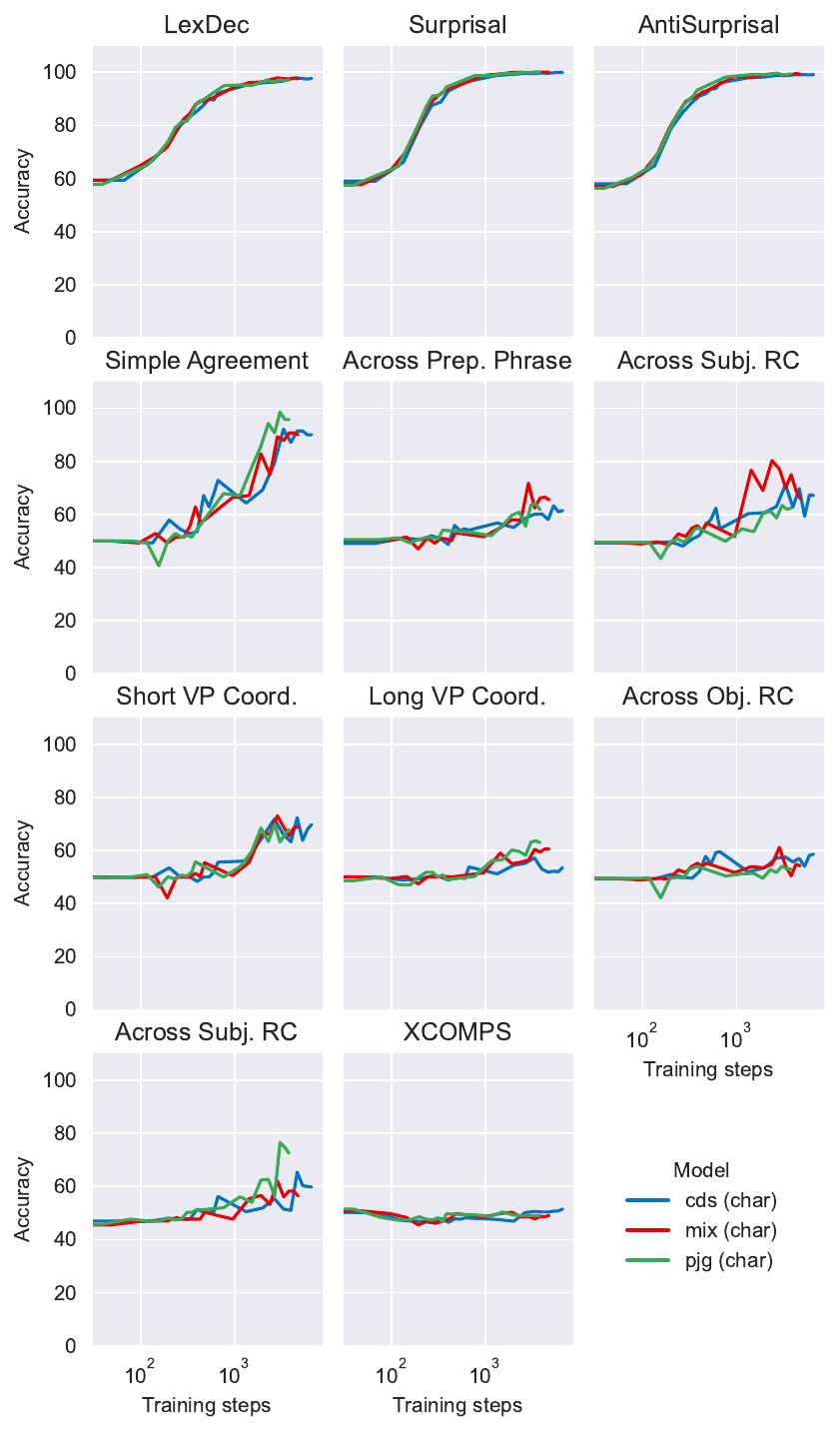}
  \vspace{-0.1cm}
  \caption{Trajectories for character models}
  \label{fig:curves-char}
\end{subfigure}%
\begin{subfigure}{.5\textwidth}
  \centering
  \includegraphics[width=0.9\linewidth]{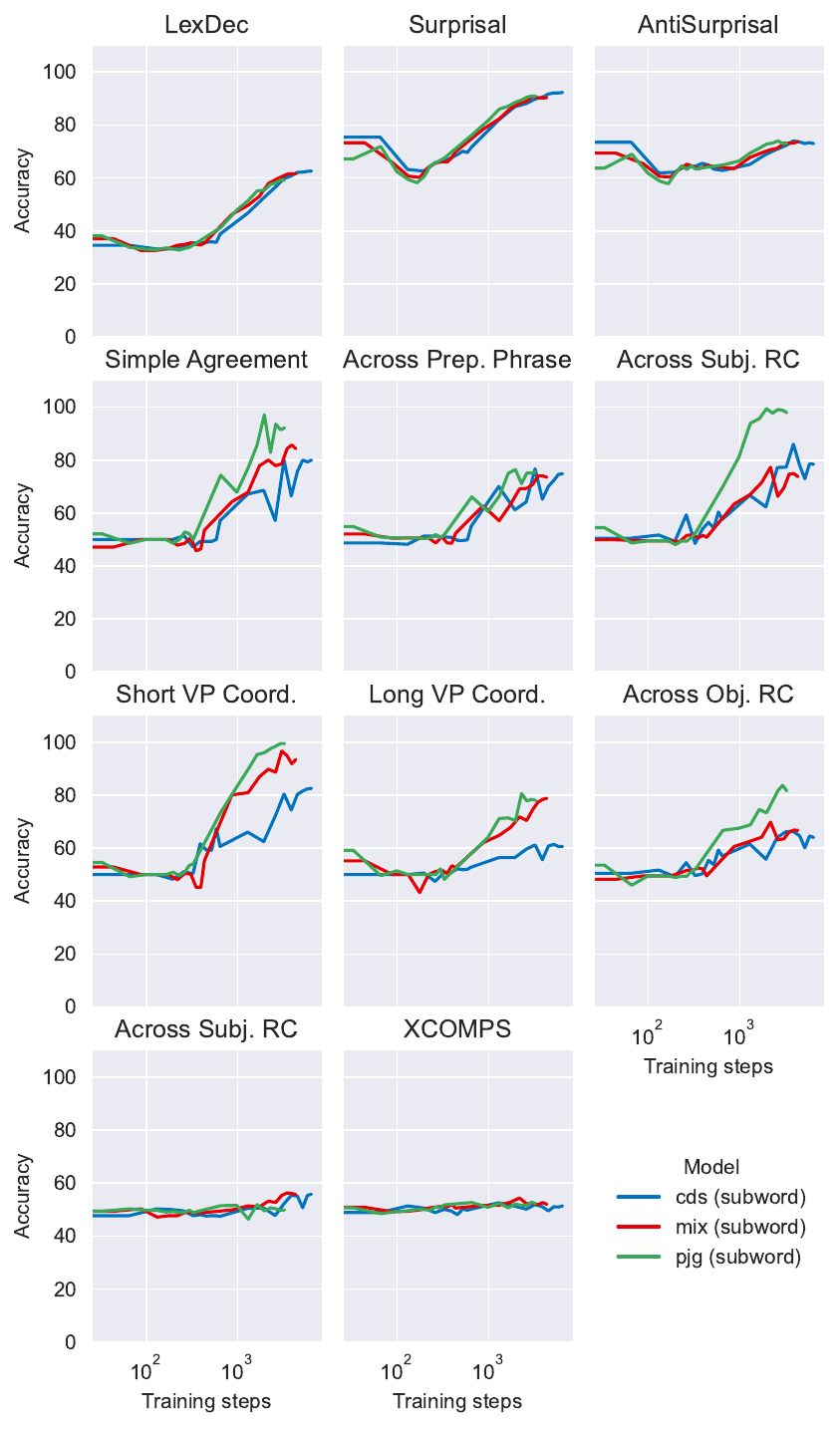}
  \vspace{-0.1cm}
  \caption{Trajectories for subword models}
  \label{fig:curves-bpe}
\end{subfigure}
\caption{Learning trajectories for all minimal pair benchmarks}
\vspace{-0.2cm}
\label{fig:learning-curves}
\end{figure*}

\section{Results}

\subsection{MP probing}
Table \ref{tab:results} shows model-wise accuracies for all minimal pair sets after training for one epoch. For the word-level evaluations, accuracy scores are generally high. Across all tasks, the character models perform with almost perfect accuracy. No effect of the constructional composition of the training data is identifiable here. For the subword models, this is not true. Here, the model trained on more questions/fragments and less complex utterances (\texttt{cds}) outperforms the model that approximates written language on the construction level (\texttt{pjg}). The improvements range from 1\% for anti-surprisal to 2-3\% on lexical decision. Interestingly, the very large ablation model (Llama 3.2 1B) performs the worst on isolated lexical decision, but reaches high scores in the surprisal setting.

For the syntactic tests, the picture is more nuanced. Generally speaking, all our models learn to distinguish most types of grammatical and ungrammatical sentences involving agreement phenomena. 
The best scores are achieved on more simplistic phenomena like simple agreement or coordination with short verb phrases. Agreement phenomena that involve longer dependencies and distracting nouns, e.g. within and across relative clauses, are the hardest to learn. For the character models, the \texttt{cds} model outperforms the others on three out of seven tests, including both ``across subj./obj. relative clause'' conditions. For three other tests, the \texttt{pjg} model wins out, whereas the \texttt{mix} model achieves the highest scores on only one test (agreement across prepositional phrases). It should be noted, that for most phenomena, the character models do perform well above chance (by a margin of 10--20\%), but still frequently make errors. The subword models show a somewhat different picture, with scores being generally higher and approximating perfect performance on 3/7 phenomena. Regarding construction distributions, the \texttt{pjg} model wins in five categories, whereas \texttt{cds} and \texttt{mix} only achieve best scores in one each. Here, the 1B-parameter Llama model outperforms our BabyLMs on 5/7 phenomena.

The scores on XCOMPS reveal that our small models do not reliably learn the conceptual knowledge underlying the included minimal pairs. Scores revolve around the chance baseline, with subword models performing slightly better than character models for 2/3 data mixtures. Nonetheless, these scores are also not considerably worse than the performance of our ablation model (58.9\%).

\subsection{Learning trajectories}

Figure \ref{fig:learning-curves} shows the learning trajectories of our models across one training epoch. As there are no intermediate checkpoints available for the 1B-parameter ablation model, we only report trajectories for our self-trained models. In line with best practices in ML \cite{viering2023shape}, we log-scale the x-axis in our plots. This allows us to also trace early learning in more detail.

For our character models, word-level learning happens rapidly in an S-shaped curve. No differences are visible between the datasets, performance improvements align almost perfectly. For the subword models, the learning processes are not as nicely monotonically improving. Rather, the learning trajectories show a dip early in training, which then later on recovers to fairly good accuracy scores. Interestingly, despite differences in final scores, the improvements across models trained on quite different datasets still align with regard to turning and takeoff points.

This pattern is also confirmed by the learning trajectories for the syntactic phenomena. While the \texttt{pjg} models trained on more complex utterances frequently reach the highest final scores, it is remarkable to see how the improvements for all models seem to happen in parallel. The global shape of the trajectory is the same for all syntactic tests, regardless of the construction distribution. For example, the learning curve for simple agreement is steeper for the \texttt{pjg} models once learning has started, but take-off points are neatly aligned. These take-off points are pushed back by the individual paradigms' complexities --- simple agreement and short VP coordination begin to improve earlier than MPs containing RCs. Finally, it is interesting to note that for the character models, word-level learning consistently stabilizes before syntactic learning, whereas both processes seem to happen concurrently in subword models (mirroring findings for English, cf. \citealp{bunzeck2025subword}). As our models do not learn to distinguish the semantic minimal pairs, the corresponding learning curves remain flat and performance differences are likely due to chance.

\section{Discussion}

This paper set out to investigate whether the constructional profile of CDS, which is shaped in a way to support the acquisition of \textit{functional} language competence, actually influences LMs' \textit{formal} language learning, and whether its relative lack of complex sentences and canonical SV(X) utterances makes it less useful training data, or too ``impoverished'' for meaningful formal learning to happen. The results of our utterance-level corpus analysis for German align with earlier findings on CDS and book language for English \cite{cameron-faulkner2003construction,cameron-faulkner2013comparison,bunzeck2024richness} and Irish \cite{cameron-faulkner2011form}, adding to the growing evidence that this linguistic distribution is fairly universal, at least in WEIRD societies \cite{henrich2024weird}.


From a language modeling perspective, the constructional profile of training data is \textit{not} overly important for the resulting performance on linguistic benchmarks. Rather, starting/turning points of the resulting learning trajectories are mostly determined by the respective amount of training steps. Despite models trained with more complex input resulting in slightly better performance, they do not begin to learn earlier. Global learning trajectories are extremely similar, only the local magnitude differs between different constructional setups. This provides further evidence that LMs based on the Transformer architecture \cite{vaswani2017attention} not only memorize language from their training data, but generalize to the underlying patterns. The same holds true or word-level learning processes such as lexical decision or (anti)surprisal tests, where data with more fragments and questions even seems to be rather beneficial. Furthermore, the comparison of our results to the Llama 3.2 1B model shows that rather high scores are already attainable with small models and little data (only on long VP-coordination do our models underperform).  

What does this now mean for theories of language acquisition? This study was inspired by findings of construction-based corpus analyses \cite{cameron-faulkner2003construction,cameron-faulkner2011form,bunzeck2024richness}, which argue that the specific constructional profile of CDS is beneficial to acquisition. Of course, LMs and minimal pair evaluations do not directly correspond to the learning processes in humans and we cannot make causal claims about them. Yet, our methodology can provide evidence as to what kinds of input data is beneficial to a purely statistical learner (that does not even tap into the functional side of language, cf. \citealp{mahowald2024dissociating}), an abstraction that is highly relevant to usage-based theories \cite{ambridge2015ubiquity}. On a formal level, there seem to be comparatively little disadvantages for models trained on less ``complex'' or somewhat impoverished data. Despite more complex data leading to slightly better benchmark scores, the learning trajectories remain largely unaffected (although somewhat erratic, cf. \citealp{bunzeck2024fifty}). What really shapes the learning process in our LMs is the amount of input, not its formal complexity (similar to findings for children by \citealp{huttenlocher1991early,rowe2012longitudinal}). An increase in appropriate construction types for child-rearing (like questions, imperatives, or fragments) does not hinder formal learning (if only reduce its magnitude slightly). As CLAMS only focuses on subject-verb agreement in canonical SV(X)-sentences, it is rather surprising that the much higher amount of questions in the \texttt{cds} dataset does not negatively affect performance, although the subjects' and predicates' positions are switched in German yes/no-questions. Conversely,  the \texttt{cds} dataset even enables word-level learning to converge to a better end state. This also aligns with a broader trend found in language acquisition studies --- the complexity and quality of input can indeed predict later language skills \cite{noble2020impact,alroqi2023association}, but the ground level is always extremely high already: being a competent user of the language itself. Furthermore, quality varies with many more extralinguistic factors like the number of siblings \cite{laing2024analyzing} or cultural factors \cite{bergelson2023everyday,bunce2024crosslinguistic}.

\section{Conclusion}

Our findings add to the growing body of research on BabyLMs \cite{warstadt2023findings,hu2024findings}. Similarly to English models, our German BabyLMs only need little data --- the \texttt{cds} dataset contains approx. 820,000 sentences, and given the estimation by \citet{cameron-faulkner2003construction} that children hear around 7,000 utterances per day, our data approximates the number of utterances heard over only 120 days --- to learn a fair amount of syntax and almost impeccable lexical knowledge, with trajectories mirroring those of English models \cite{bunzeck2025subword}. We hope that our dataset enables other scholars to carry out experiments with developmentally plausible LMs beyond the dominating English LMs, and that our data provides inspiration to those compiling BabyLM corpora for other languages.



\section*{Limitations}

Our study is limited by data availability. Creating a full-fledged 100M-token BabyLM dataset with \textit{only} child-directed speech or other explicitly child-directed materials is currently out of question, as neither CHILDES nor other sources contain even remotely enough data for languages other than English. To reach higher token counts, padding with larger data sets, e.g. more tokens from the OpenSubtitles dataset, would be necessary. Principally, synthetic corpora like the TinyStories dataset \cite{eldan2023tinystories}, which contains children's stories generated by GPT-3 or TinyDialogues by \citet{feng2024childdirected} would provide an unlimited source of training data. However, our inspection of their generated dialogues yielded that they drastically underestimate the high numbers of grammatical fragments, questions and short SV(X)-utterances in real-world data. Similarly, there are little to no evaluation sets specifically aimed at German, beyond those that we included/creates ourselves, especially on the syntactic level. Only very recently, evaluation datasets like the massively multilingual MultiBLiMP have begun to fill this gap \cite{jumelet2025multiblimp}. Also, such minimal pair datasets are principally at odds with the usage-based, constructionist view on language development, because they are grounded in the Generativist notion of defining rules that can determine whether an utterance belongs to a language or not, whereas usage-based linguistics has adopted a network-based, associative model of linguistic knowledge \cite{diessel2019grammar,diessel2023constructicon}. As of late, these developments have begun to make their way into the broader LM evaluation landscape
\cite{weissweiler2025linguistic}, and novel evaluation methods like measuring affinities between lexical items and testing if different constructions manifest from them \cite{rozner2025constructions,rozner2025babylms} provide promising future research avenues.

Moreover, actual developmental plausibility also hinges on the inclusion of other modalities. For audio data, there are few CHILDES subcorpora and other corpora that contain phonetic information \cite{lavechin2023babyslm}, but larger models need to be trained on more data, e.g. audiobooks \cite{lavechin2025simulating}. A middle ground is training on textual phonetic transcriptions generated from raw text, e.g. for the BabyLM data \cite{goriely2024babble,bunzeck2025small,goriely2025ipachildes}. More recently, also video recordings from infant-mounted cameras have been used to train on combined visual and auditory input modalities \cite{wang2023finding,vong2024grounded,long2024babyview}. The inclusion of such data could help to disentangle learning processes further.

\section*{Ethical considerations}

Given the nature of this work, there are no specific ethical concerns to address. 
However, we would like to stress that, of course, BabyLMs are not supposed to simulate real babies, but rather to instantiate abstractions, or \textit{models} in the original scientific sense, of the distributional, frequency-driven aspects of their learning capacity. All claims regarding their implications for language development in the real world should be understood in this context, which we also attempted to explicate by distinguishing functional and formal aspects of learning.

\section*{Acknowledgments}
This research has been funded by the Deutsche Forschungsgemeinschaft (DFG, German Research Foundation) --- CRC-1646, project number 512393437, project A02.

\bibliography{bastian,dd}

\appendix
\newpage
\section{Excluded corpora}

Several corpora that are --- in principal --- available for German were excluded from our analysis. The Folk corpus \cite{reineke2023forschungs} and the Simple German corpus \cite{jach2024korpus} are not available under any open licenses, while the data in other German reference corpora \cite{kupietz2010german} are not available in their entirety but can only be queried through web interfaces. Finally, Homebank features day-long audio recordings of children and their surroundings/inputs \cite{vandam2016homebank}, but without any written transcriptions.

\section{Data cleaning} \label{app:data-cleaning}

In line with best practices in language modeling, we extensively clean and normalize our data.

\paragraph{All subcorpora} We replaced all local variants of single/double quotation marks with either \verb|' '| or {\textquotedbl\ \textquotedbl}. We further reduced multiple superfluous whitespace and newlines to singular whitespaces.

\paragraph{Talkbank data} For the data sourced from talkbank (i.e. the CHILDES corpora and CallHome), we remove all mark-up and additional info on false starts, hesitations, implicit completions or other explanations. Furthermore, we also remove all empty utterances and those containing \texttt{xxx} or \texttt{yyy}, placeholder symbols for personally identifiable information.

\paragraph{Project Gutenberg} For the Project Gutenberg data, we excluded all lines with more than 6 consecutive whitespaces, as these always turned out to be title pages, index pages, etc., which contain no useful language data. Additionally, we removed all textual data in square brackets, which almost always corresponded to pointers to pictures which are not found in text-only version, or additional explanations by the volunteers who digitized the respective books.

\paragraph{OpenSubtitles} For the OpenSubtitles data, we removed all text in parentheses, which corresponds to speaker information. Also, we removed sentence-initial dashes (\texttt{-}) which were sometimes added. We also amended OCR errors (like mangled uppercase I and lowercase l) as far as possible.

\paragraph{Fluter} For the data sourced from the Fluter magazine, we removed all lines containing additional metatextual data, like author info and image credits, before pre-training.

\onecolumn

\section{Exact construction proportions} \label{app:exact-cxn-proportions}

Table \ref{tab:cxn-proportions} shows the exact construction proportions for all of our subcorpora. This data underlies the visualization in Figure \ref{fig:cxn-proportions}.

\begin{table*}[ht!]
\scriptsize
\centering
\begin{tabular}{@{}l|rrrrrrrrrrr@{}}
\toprule
Construction & Proj. Gut. & Dreamb. & Fluter & News & Wikib. & Klex. & Mini-Klex. & OpenSub. & CallHome & Child speech & CDS \\ \midrule
FRA & 7.8\% & 6.3\% & 6.2\% & 4.0\% & 11.6\% & 6.3\% & 2.5\% & 24.1\% & 37.0\% & 55.1\% & 24.5\% \\
QWH & 1.9\% & 0.3\% & 2.6\% & 1.4\% & 0.5\% & 2.9\% & <0.1\% & 7.3\% & 2.1\% & 3.5\% & 8.8\% \\
QYN & 3.7\% & 0.7\% & 2.8\% & 1.6\% & 0.5\% & 0.4\% & <0.1\% & 10.9\% & 6.9\% & 4.7\% & 20.7\% \\
COP & 4.6\% & 7.1\% & 7.7\% & 7.4\% & 10.9\% & 13.2\% & 21.4\% & 9.7\% & 10.7\% & 5.7\% & 8.1\% \\
IMP & 1.5\% & 0.1\% & 0.2\% & 0.1\% & 0.3\% & <0.1\% & <0.1\% & 4.6\% & 0.4\% & 2.0\% & 4.5\% \\
SPI & 7.5\% & 9.2\% & 9.7\% & 13.7\% & 9.5\% & 13.9\% & 19.9\% & 9.9\% & 8.8\% & 11.5\% & 10.1\% \\
SPT & 10.5\% & 14.5\% & 18.7\% & 25.7\% & 24.1\% & 28.1\% & 37.2\% & 18.0\% & 14.1\% & 11.9\% & 12.3\% \\
COM & 62.5\% & 61.8\% & 52.2\% & 46.1\% & 42.7\% & 35.2\% & 18.9\% & 15.4\% & 20.0\% & 5.7\% & 11.0\% \\ \bottomrule
\end{tabular}
\caption{Exact proportions of constructions for all subcorpora}
\label{tab:cxn-proportions}
\end{table*}

\section{Model hyperparameters and training details} \label{app:loss-curves}

Our models share a hidden/intermediate/embedding size of 256, 8 hidden layers and attentions heads, and a context length of 128. For the character models, the vocabulary consists of all printable ASCII characters and characters used in written German (üäöß and their uppercase variants), amounting to a vocab. size of 110 and 3,730,688 parameters. For the subword models, we train a BPE tokenizer \cite{gage1994new} with a vocab. size of 8,000 and add two special tokens (BOS, EOS/PAD), resulting in 8,002 vocab. tokens and 7,771,392 parameters. Model training takes approx. 2h on a MacBook Pro with an Apple M2 Pro CPU/GPU.

\begin{figure*}[ht!]
\centering
\includegraphics[width=0.8\linewidth]{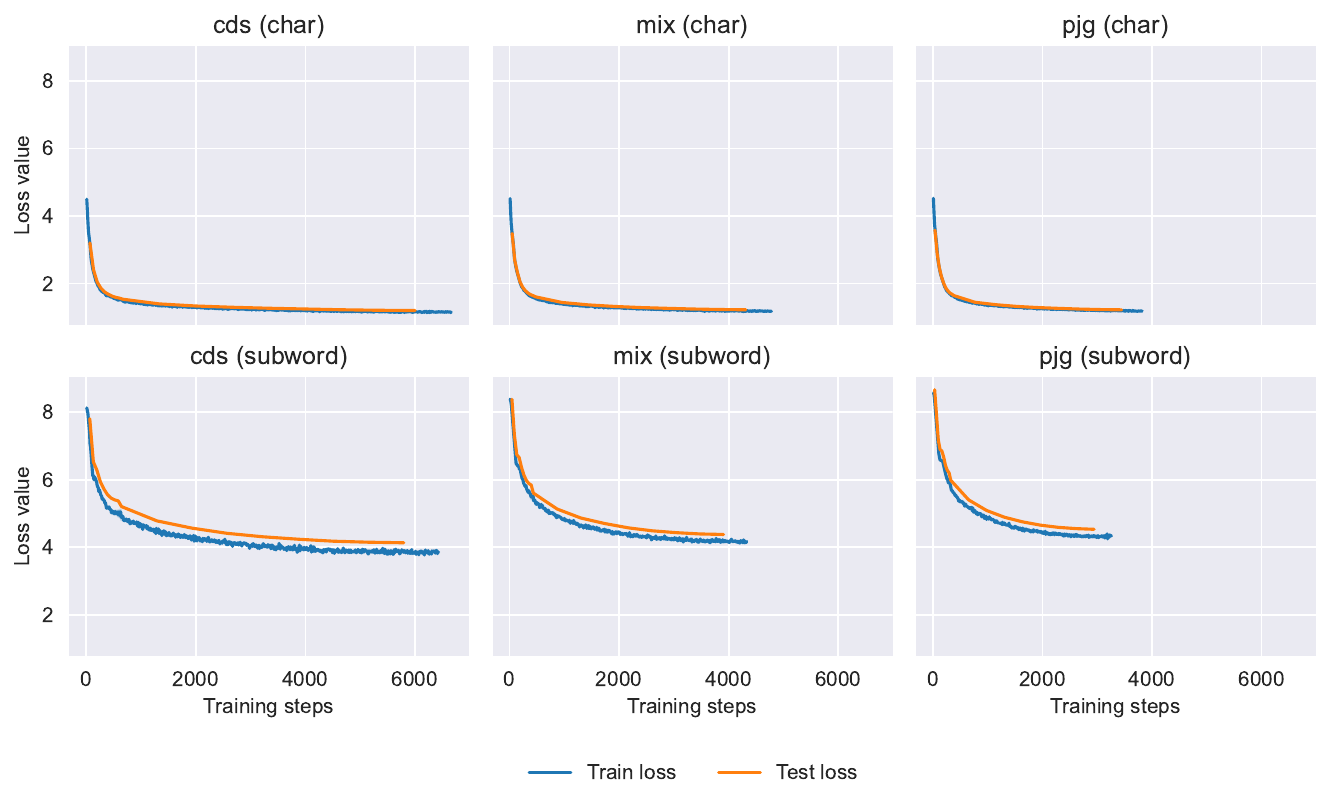}
\caption{Loss curves for our self-trained character and subword models}
\label{fig:loss}
\end{figure*}

We reproduce the training and test loss curves for our models in Figure \ref{fig:loss}. For the test loss, we evaluated perplexity on a held-out, randomly sampled portion of each individual training corpus. We find no principal differences in loss development, although the character models and models trained on the \texttt{cds} data seem to converge the fastest. As the similar curves for train and test loss indicate, all models succeed in optimizing for their next-token prediction goal. It should be noted that due to longer/shorter sequences in the different data mixtures and our choice of padding to the maximum sequence length, some models are trained for more \textit{steps}, although the number of \textit{lexical tokens} remains the same.

\newpage
\section{Repeated training runs} \label{app:repeated-runs}

\begin{figure*}[ht!]
\centering
\begin{subfigure}{.5\textwidth}
  \centering
  \includegraphics[width=0.9\linewidth]{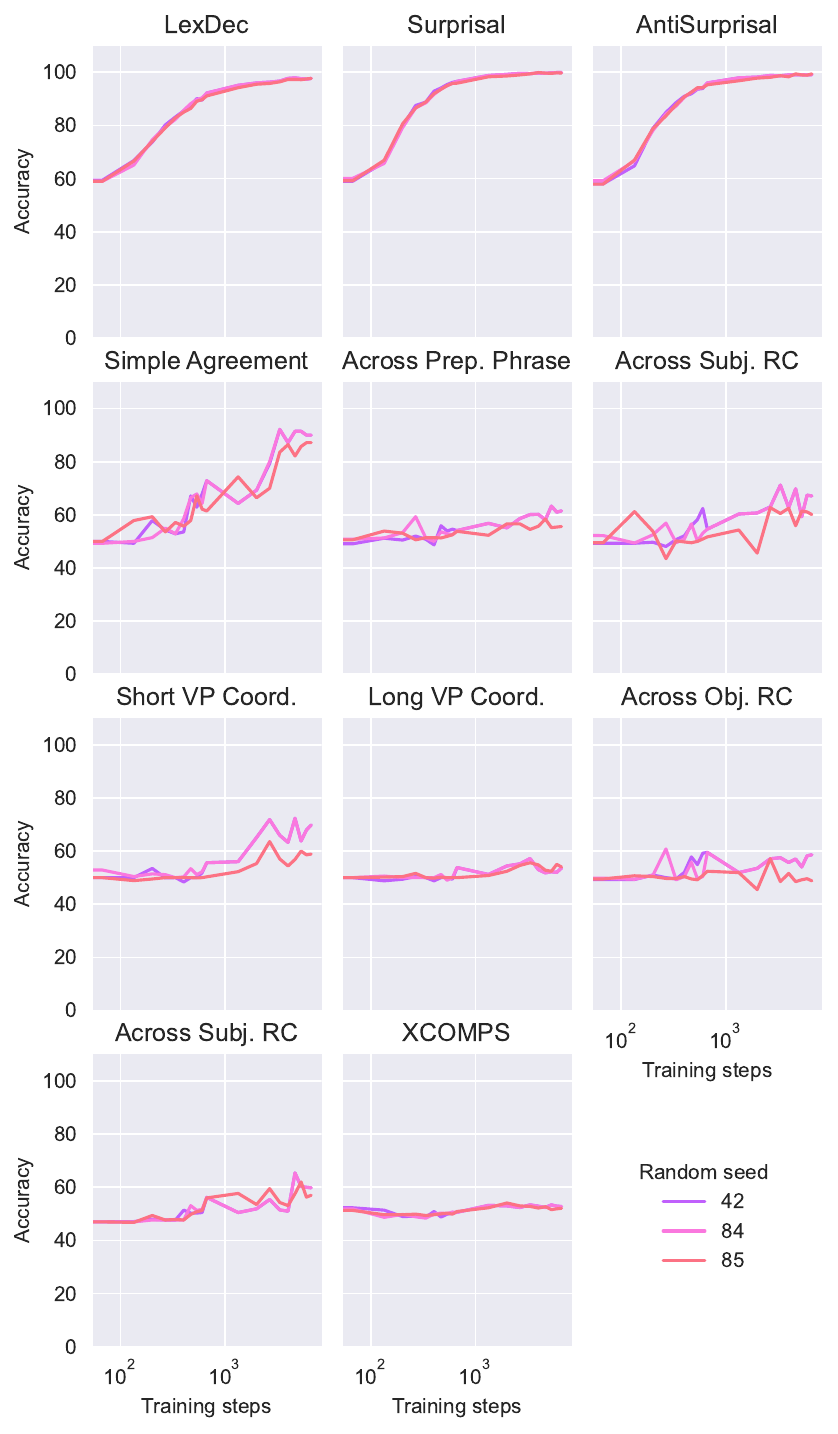}
  \caption{Trajectories for different random initializations}
  \label{fig:curves-repl}
\end{subfigure}%
\begin{subfigure}{.5\textwidth}
  \centering
  \includegraphics[width=0.9\linewidth]{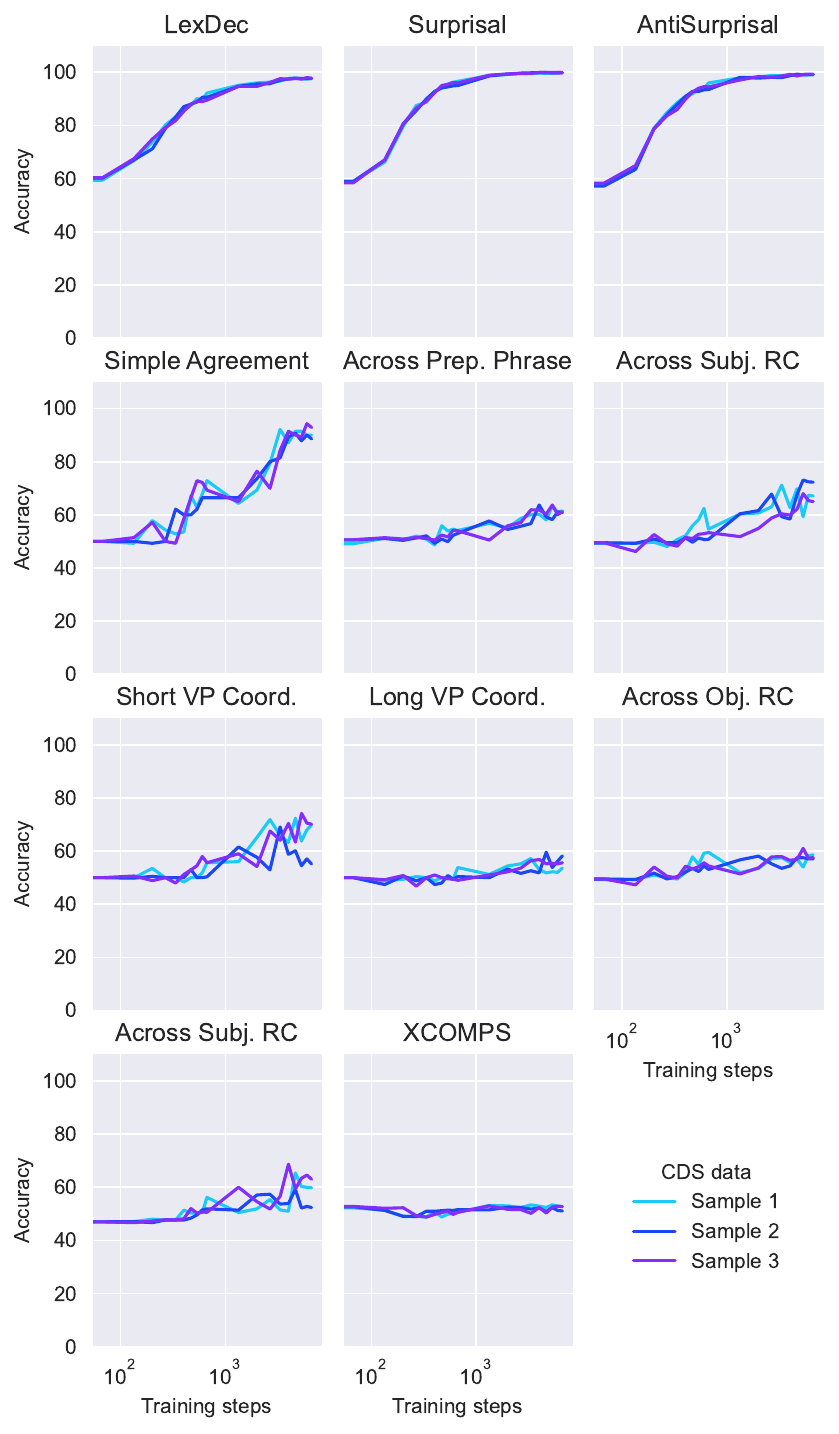}
  \caption{Trajectories for different samples of \texttt{cds} data}
  \label{fig:curves-samp}
\end{subfigure}
\caption{Learning trajectories for our comparison models}
\label{fig:curves-ablations}
\end{figure*}

A common criticism towards the BabyLM paradigm is the purported effect of training noise on model performance, which is hard to disentangle from real training data effects. While training and evaluating multiple random seeds for all our models would be too costly, we repeated two additional training runs for the character-level \texttt{cds} model with different random initializations (learning trajectories in Figure \ref{fig:curves-repl}) and two additional training runs where we re-sampled the \texttt{cds} dataset from our whole corpus with the exact same construction composition, but different content (learning curves in Figure \ref{fig:curves-samp}). In both cases, the learning trajectories do not differ tremendously. For the word-level phenomena (LexDec, Surprisal, AntiSurprisal), the curves overlap almost perfectly. For the syntax phenomena, we can see some variation and oscillation in the curves, but the trajectories still remain extremely similar (and do not differ in their steepness, the main effect that we see in Figure \ref{fig:learning-curves} between the datasets with different construction compositions).

\end{document}